\documentclass[letterpaper, 10 pt, conference]{ieeeconf}  

\IEEEoverridecommandlockouts                              

\usepackage{graphics} 
\usepackage{graphicx} 
\usepackage{amsmath} 
\usepackage[noend]{algpseudocode}
\usepackage{algorithmicx,algorithm}
\usepackage{cite}

\usepackage{multirow}
\usepackage{hyperref}
\usepackage{mathrsfs}
\usepackage{bbm}
\usepackage{color}
\usepackage{tabularx}
\usepackage{threeparttable}
\usepackage{amssymb}
\usepackage{xspace}
\usepackage{float}
\usepackage{booktabs}
\usepackage{subcaption}
\usepackage{mwe}
\usepackage{threeparttable}  

\usepackage{amssymb}
\usepackage{bbding}
\usepackage{algorithm}
\usepackage{caption}
\usepackage[misc]{ifsym}
\title{\LARGE \bf
Density-based Curriculum for Multi-goal Reinforcement Learning with Sparse Rewards
}

\author{ Deyu Yang, Hanbo Zhang, Xuguang Lan*, Jishiyu Ding

\thanks{
Corresponding Author: Xuguang Lan.
Xuguang Lan is with the Institute of Artificial Intelligence and Robotics, Xi'an Jiaotong University, No.28 Xianning Road, Xi'an, Shaanxi, China.
        {\tt\small xglan@mail.xjtu.edu.cn}
        }
  \thanks{This work was supported in part by NSFC under grant No.6212500145, No.62088102, No.61973246, No.91748208, Shaanxi Project under grant No.2018ZDCXLGY0607, and the program of the Ministry of Education.}
}

\begin{document}

\maketitle
\thispagestyle{empty}
\pagestyle{empty}

\begin{abstract}
Multi-goal reinforcement learning (RL) aims to qualify the agent to accomplish multi-goal tasks, which is of great importance in learning scalable robotic manipulation skills. 
However, reward engineering always requires strenuous efforts in multi-goal RL.
Moreover, it will introduce inevitable bias causing the suboptimality of the final policy.
The sparse reward provides a simple yet efficient way to overcome such limits.
Nevertheless, it harms the exploration efficiency and even hinders the policy from convergence.
In this paper, we propose a density-based curriculum learning method for efficient exploration with sparse rewards and better generalization to desired goal distribution.
Intuitively, our method encourages the robot to gradually broaden the frontier of its ability along the directions to cover the entire desired goal space as much and quickly as possible.
To further improve data efficiency and generality, we augment the goals and transitions within the allowed region during training. 
Finally, We evaluate our method on diversified variants of benchmark manipulation tasks that are challenging for existing methods. 
Empirical results show that our method outperforms the state-of-the-art baselines in terms of both data efficiency and success rate.
\end{abstract}

\section{Introduction}
Reinforcement learning (RL) has attracted widespread 
attention in recent years with inspiring results in video games~\cite{DQN,deepLFgame} and complex robot manipulation tasks~\cite{EndTend,leveragdemonstration,demonsRL}. Despite its notable success, RL always needs arduous reward engineering, which might bias the learning and limit the scalability of the learned policies. Therefore, it's more natural and convenient to use sparse reward settings in manipulation tasks, i.e., the robot will be awarded only when it reaches the final goal. However, the sparse reward makes it extremely hard to explore in the initial stage, resulting in few useful learning signals, and even hindering the agent from convergence.

Andrychowicz et al. \cite{HER} proposed a Hindsight Experience Replay (HER) to tackle  sparse-reward RL in multi-goal robot manipulation tasks. By replacing desired goals with the randomly selected achieved goals, namely hindsight goals, of the past experiences, an agent can get much more positive reward signals and learn to master skills of reaching handy goals. However, this mechanism does not provide the guarantee of accomplishment of original tasks. The agent is able to generalize to finishing original tasks only when the distribution of hindsight goals is close enough to those originally desired goals. When there's little or no overlap between desired and hindsight goal distributions, the originally desired goals are too hard for HER agent to achieve with random hindsight goal selection. 

\begin{figure}
	\centering
    \setlength{\belowcaptionskip}{-0.5cm} 
    \includegraphics[width=0.48\textwidth]{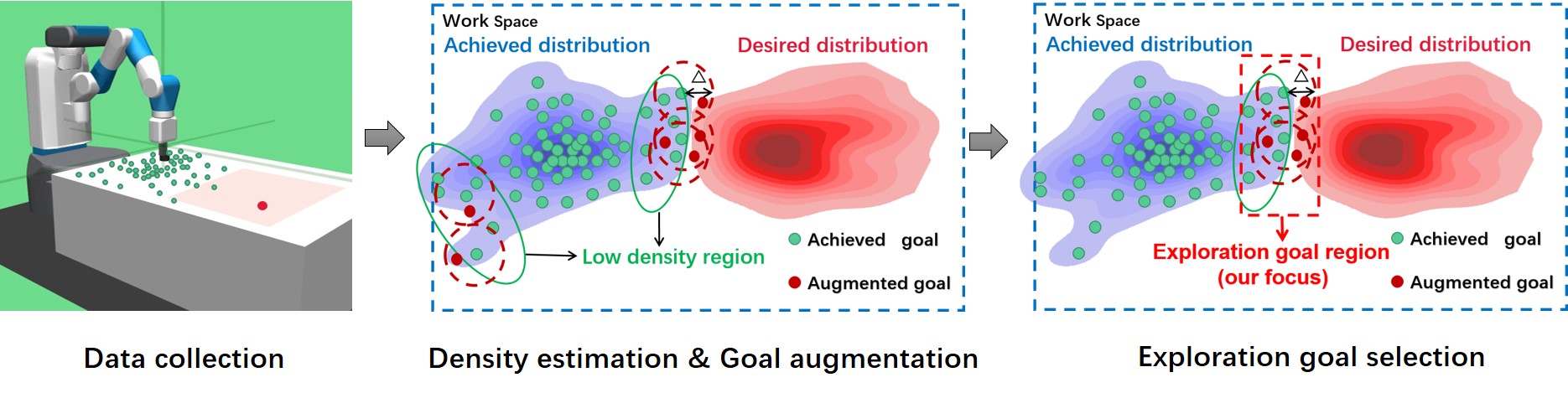}
    \caption{Illustration of goal selection. We utilize density estimation and goal augmentation to select goals from the exploration goal region. }
    \label{illustration}
    \vspace{-0.15cm}
\end{figure}

In this paper, we propose to use a curriculum for the efficient exploration, which gradually expands the frontier of the reachable range along the directions to cover the entire originally desired goal space. To be specific, 
as shown in Fig. \ref{illustration}, we firstly evaluate achieved goal distribution using Kernel Density Estimation (KDE) \cite{KDEstimate} and select a set of goals in the low-density region of the distribution. Then we rank the selected goals by measuring the element-wise entropy computed by the estimated desired goal distribution. Exploring and exploiting these goals (called exploration goals) makes the agent push the frontier of achieved goals forward and simultaneously generalize better to the desired goal space. Moreover, we propose goal and transition augmentation to achieve higher data efficiency and better generalization.

To summarize, our main contributions of the paper are : 
\begin{itemize}
    \item We propose a novel density-based curriculum learning method to select hindsight goals for hindsight reinforcement learning, which helps guide the agent efficiently explore towards the desired goal distribution.
    \item We introduce two simple but effective data augmentation methods: goal augmentation and transition augmentation to help further improve the data efficiency and generality.
    \item We evaluate our method on several challenging variants of robotic manipulation tasks and demonstrate that our method achieves state-of-the-art performance compared to other baselines.
\end{itemize}

\section{Related work}

Our work focuses on the problem of how to design a curriculum to efficiently guide the agent to explore the informative space during reinforcement learning with sparse rewards. Therefore, in this section, we will review the key related works on these topics.

\subsection{Exploration in Sparse-reward RL}
\label{exploration}
Informative and effective explorations are essential for solving RL tasks. 
However, for sparse-reward RL, traditional exploration strategies, including $\epsilon$-greed~\cite{Qlearning},  Ornstein–Uhlenbeck noise~\cite{DDPG}, maximum entropy explore in action space~\cite{SAC}, and noise adding in parameter space~\cite{ParameterNoise, NoisyNetwork} cannot work well since they cannot help the agent to receive enough positive rewards.

Recent works impel agents to discover novel states by intrinsic motivation.
They can be mainly divided into three categories: count-based methods, curiosity-driven methods, and entropy-based methods.
Count-based methods estimate visit counts and turn them into intrinsic bonuses to help agents reduce uncertainty~\cite{modelbasedEstimate,modelbasedEstimate2,CountBasedDensityEsti,hashexplore}.
Such methods are not suitable for tasks with large state space like robotic manipulation.
The curiosity-driven method often utilizes prediction errors in the learned feature space to provide intrinsic reward~\cite{ICM,PredicterrorExploration,RandomNetwork,EMI,made,ex2}. 
However, these shaped rewards may cause the learning process detoured and unstable~\cite{GenerativeEA}.
The entropy-based method aims to maximize information gain~\cite{Pitis2020MaximumEG,VIME,AKL,modelBaseAE}, state entropy~\cite{maximumentropy,StateMarginalMatch,StateEMwithRandomEncoder}, or mutual information between the agent's action and the next states~\cite{EMI,music} to encourage agents to discover novel states or explore unexplored regions of the environment.

In our work, we focus on setting curriculum-based goals step by step to guide the agent explore the environment efficiently. 
The most related work to ours is OMEGA~\cite{Pitis2020MaximumEG}, which selects achieved goals in sparsely explored areas to maximize entropy gain to push the frontier of achieved goals forward. However, OMEGA usually turns to unsafe goals (e.g. below the table) pursuit at the early exploration because of the instability of the mixture distribution, which may hinder the learning process or even cause irrevocable damages to the environment. By contrast, our work utilizes estimated desired goal distribution to constraint novel goal states and use goal augmentation to quickly expand the boundary of achieved goals, thus making the exploration more efficient and safe.

\subsection{Curriculum Learning}
\label{curriculum}
Our work is also highly related to curriculum learning (CL)~\cite{bengio2009curriculum}, which  has been introduced to deep RL to guide data collection and exploitation~\cite{portelas2020curriculum}.
During data collection, CL defines a set of auxiliary tasks to be used as stepping stones towards the main task. 
For example, CL could be used to generate a sequence of sub-goals~\cite{hindisightplanner,ImaginSubgoal,DataEfficientHR,followobject} to guide the agent to finish the original task step by step.
In multi-goal reinforcement learning, CL could be also applied to guide the goal selection~\cite{goalgan, plangan, tomar2019mamic, skew-fit, HGG, VDS}.
Particularly, hindsight goal generation (HGG) \cite{HGG} is a state-of-the-art method in the area of exploration algorithms in sparse-reward RL, which optimizes desired goals by solving the Wasserstein Barycenter problem.
CL can also be used in data exploitation by performing transition prioritization and transition modification~\cite{MEP, energybasedER, CHER}. 
For example, CHER \cite{CHER} adaptively selects the failed experiences for replay according to the proximity to original goals and the curiosity of exploration. However, the metric to select goals is hand-designed and influenced by hyperparameters. Besides, the selection mechanism~\cite{laziersample} of CHER is quite time-consuming. 
By contrast, our work utilizes a more generalizable KDE algorithm without increasing computational complexity.

\section{Background}
\subsection{Goal-conditioned Reinforcement Learning}
Compared with the standard RL which is formulated by Markov Decision Process (MDP) represented as a tuple $(\mathcal{S}, \mathcal{A}, \mathcal{P}, r,\gamma, \rho_{0})$, Goal-conditioned Reinforcement Learning (GCRL) arguments it with an additional set of goals $\mathcal{G}$. Formally, the Goal-conditioned MDP is denoted as a tuple $(\mathcal{S}, \mathcal{A}, \mathcal{G}, \mathcal{P}, r_{g}, \gamma, \rho_{0})$, where $\mathcal{S}$ and $\mathcal{A}$ are set of states and actions, respectively, $\mathcal{P}=p(s_{t+1}|s_t,a_t)$ defines the transition dynamics of the environment. $\mathcal{G}\in \mathbb{R}^{d_g}$  is the goal space and $r_{g}:\mathcal{S} \times \mathcal{A} \times \mathcal{G} \rightarrow \mathbb{R}$ defines the reward function that describes how close the agent is to the desired goals. $\gamma \in [0,1]$ is a discount factor and $\rho_{0}=p(s_{0})$ is the initial state distribution. Specifically, in the sparse reward setting, the reward function has the following form:
\begin{equation}\label{reward_fun}
r_{g}\left(s_{t}, a_{t}, g\right)=\left\{\begin{aligned}
0, & \left\|\phi\left(s_{t}\right)-g\right\| \leq \delta \\
-1, & \text { otherwise }
\end{aligned}\right.,
\end{equation}
where $\phi: \mathcal{S} \rightarrow \mathcal{G}$ is a given function that maps state space to goal space. $\delta$ is a know distant threshold that represents the tolerance of reached goal. $||\bullet||$ is a distant metric.

\subsection{Deep Deterministic Policy Gradient}

Deep Deterministic Policy Gradient (DDPG)~\cite{DDPG} is an off-policy model-free reinforcement learning algorithm, which utilizes an actor-critic~\cite{RLIntroduction} framework and  is proposed to deal with continuous control tasks. The actor is for approximating policy $\pi:\mathcal{S} \rightarrow \mathcal{A}$ and the critic aims to learn action-value function $Q:\mathcal{S} \times \mathcal{A} \rightarrow \mathbb{R}$. 


As for multi-goal continuous control tasks, DDPG can be extended with Universal Value Function Approximators (UVFA)~\cite{UniversalVF}, whose idea is to augment the Q-function and policy with multiple goal states $g\in \mathcal{G}$. Thus the actor and critic network  additionally depend on goal state: $\pi =\pi(s,g), Q=Q(s,a,g)$.

\subsection{Hindsight Experience Replay}

Hindsight experience replay (HER)~\cite{HER} was proposed to solve the sparse reward environments in multi-goal RL.  HER  learns from failed experience using goal relabeling  technique. Given a trajectory $\tau=\left\{\left(s_{t}, a_{t}, g, r_{t}, s_{t+1}\right)\right\}_{t=1}^{T}$ of length $T$, HER alternates $g$  and $r$ in the $i$-th transition $(s_i, a_i, g, r_i, s_{i+1})$ with a future achieved goal in the same trajectory $g^{'} = \phi(s_{i+k}), 0 <{k} \leq {T-i}$ and $r^{'}=r(s_i, a_i, g^{'})$ computed by Eq. (\ref{reward_fun}). Through relabeling, transitions from a failed trajectory can gain non-negative reward, thus helping the agent's learning in sparse reward environment.
HER can be combined with any off-policy algorithms such as DDPG~\cite{DDPG}, TD3~\cite{TD3}, SAC~\cite{SAC}. In this paper, we adopt DDPG+HER framework following~\cite{HER}.


\section{Method}
\subsection{Overview}
\begin{figure}
	\centering
    \setlength{\belowcaptionskip}{-0.5cm} 
	\includegraphics[width=0.45\textwidth]{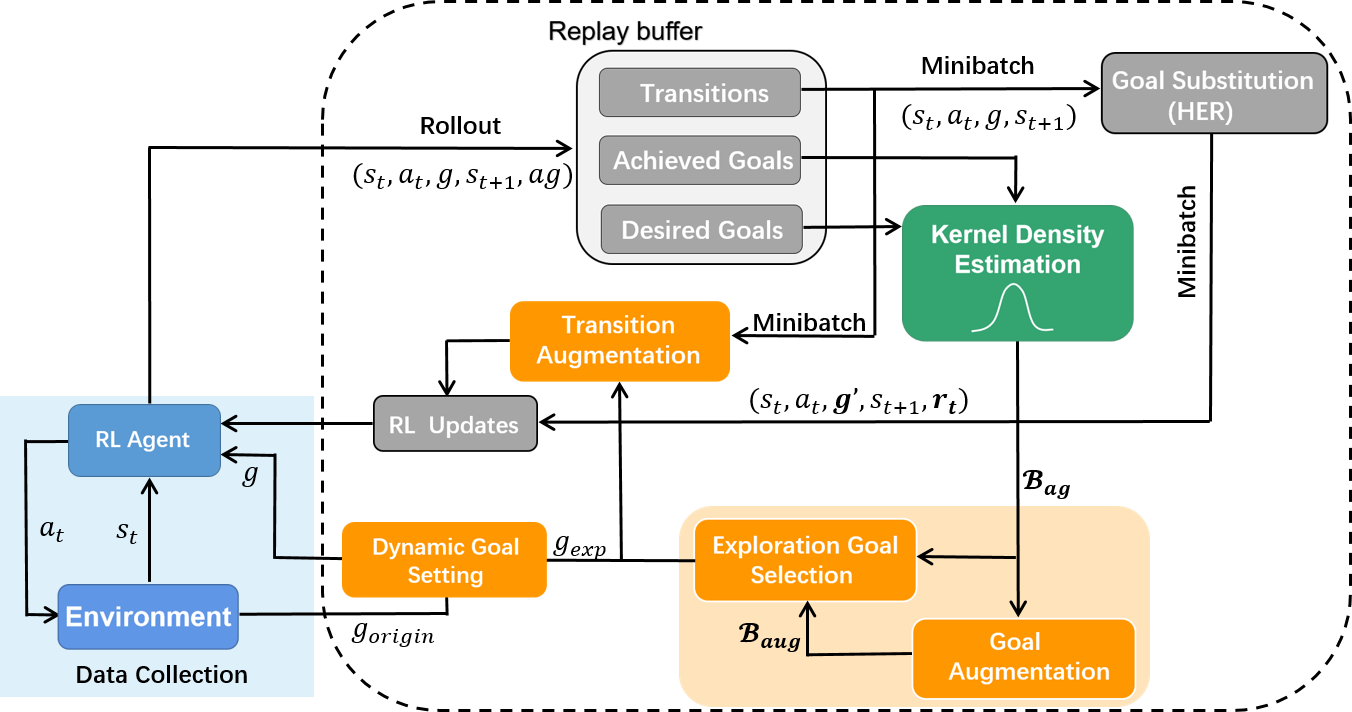}
    \caption{Framework of our proposed method. The dash box contains the main modules of our method.}
    \label{framework}
\end{figure}

The overview of our proposed method is shown in Fig. \ref{framework}. We mark the key components in orange and green colors.  
Roughly, we follow the standard off-policy RL skeleton, in which the past experiences from the interaction between the agent and the environment will be stored in the replay buffer and used in policy updates.
The key contribution of our work lies in the dashed box defining the sampling and post-process of past experiences.
To be specific, firstly, we store the achieved and desired goals into the corresponding replay buffer after trajectory data collection. Then we fit density models of achieved and desired distribution using the KDE module, and select exploration goals according to Section \ref{achived_curriculum} and \ref{desired_curriculum} . These exploration goals have two aspects of usefulness. One (Dynamic exploration goal setting module, see Section \ref{dynamic_setting}) is to replace the original goal provided by the environment with probability $\epsilon$ at the beginning of trajectory collection, the other (Transition augmentation module, see Section \ref{Augmentation}) is to relabel transition goals like HER substitution mechanism. Finally, we  mix augmented transitions with  relabeled ones that use HER \textit{future} strategy  for policy updates. 

By setting exploration goals at the beginning of every training epoch, the agent is  capable to get more guidance to explore along the reachable boundary. Moreover, by cooperating with goal and transition augmentation, the agent could achieve better generalization and performance.

\subsection{Curriculum via Achieved Goal Density}
\label{achived_curriculum}
We hope that the agent could set its own exploration goals to avoid pursuing unobtainable goals.
But which goals are more valuable and should be selected for exploration?
One notable fact in HER is that achieved goals easily get stuck in the area around the starting point during early exploration. However, learning these repeated goal states is useless to improve the agent's capability, while goal states with low reachability are more informative to extend the reachable boundary. Therefore, we focus on the low-density region of achieved goals, which is similar to the curiosity learning.

To do so, during exploration, we collect achieved goals and  store them in the replay buffer. Then we introduce the KDE algorithm to estimate their density, and thus obtain the density model of the achieved goals $M_{ag}$. For an achieved goal $ag_{i}$ in the replay buffer, its density can be computed as 
\begin{equation}
\rho_{i} = M_{ag}(ag_{i}),
\end{equation}
then the density is normalized using the following equation:
\begin{equation}
\hat{\rho}_{i}=\frac{\rho_{i}-\rho_{\min }}{\sum_{n=1}^{N}\left(\rho_{n}-\rho_{\min }\right)},
\end{equation}
where $\rho_{min}$  represents the minimum value of density, and $N$ is the size of the achieved goal buffer. After normalization, all density values are limited within the range $0 <\hat{\rho}<1$. To ensure that low-density achieved goals have higher prioritization, we use $1-\hat{\rho_{i}}$ and softmax function to calculate sample probability of $ag_{i}$ and store the value in the goal replay buffer.
\begin{equation}
p(ag_{i})=\frac{e^{1-\hat{\rho_{i}}}}{\sum_{n=1}^{N} e^{1-\hat{\rho_{n}}}}
\label{eq:achievedcurriculum}
\end{equation}
Finally, we will sample a batch of achieved goals $\mathcal{B}_{ag}=\left\{g_{i}\right\}_{i=1}^{n}$ according to sample probability defined in Eq. (\ref{eq:achievedcurriculum}).

\subsection{Curriculum via Desired Goal Density}
\label{desired_curriculum}
 Although goals in $\mathcal{B}_{ag}$ could be more informative to learn, not all these goals are safe or valuable for exploration. Actually, some goals are exactly opposite to the desired distribution or out of the safe workspace (e.g., below the table), which are useless to learn and may cause damages to the environment. To constrain the selected goals, we propose to utilize desired goal density model to further rank the selected goals. Formally, for each selected achieved goal $g_{i}$, its density probability in desired goal distribution can be calculated as follows:
\begin{equation}
p(g_{i})=M_{dg}(g_{i}),
\label{density}
\end{equation}
where $M_{dg}$ is the desired goal density learned in a similar way to $M_{ag}$. 

Goals with higher desired density will be naturally sampled more frequently and hence dominate the learning process.
However, we found that such domination will harm the final performance since the agent tends to overfit such high-density goals.
To achieve better performance, the agent should not only learn to reach high-density goals, but also be capable of relatively low-density goals. 
Therefore, learning to reach goals that lie in the uncertain areas of the desired distribution is valuable to improve the final performance. 
Here, we choose element-wise entropy to measure the uncertainty of $g_{i}$ in desired goal distribution following ${e}(g_{i})=-p(g_{i})\log p(g_{i})$, and the normalized entropy is 
\begin{equation}
\hat{e}(g_{i})=\frac{e(g_{i})}{\sum_{j=1}^{n}{e(g_{j})}}.
\label{normalize_entropy}
\end{equation}
We rank the selected achieved goals $\mathcal{B}_{ag}$ by the normalized entropy and then only preserve the top-ranking ones to form the training batch.
Intuitively, such a mechanism will impose high priority on goals at the boundary of the desired distribution, and ignore the ones with extremely low density, which are not valuable to learn, and high density, which have already been enough learned.
It helps to achieve better generalization to the whole distribution instead of only the high-density area. 


\subsection{Dynamic Exploration Goal Setting}
\label{dynamic_setting}
Our method is proposed to help the agent efficiently explore the environment especially at the beginning stage.
However, such goal relabeling techniques will inevitably introduce learning bias, i.e., it encourages the agent to finish the relabeled tasks, and sometimes prevents the agent from learning to finish the original tasks.
Therefore, we hope that the agent could explore in high efficiency with our method at the beginning, and gradually turn to learn the original tasks after enough exploration.

To achieve this, at the beginning of each episodes, we replace the original goal provided by the environment with the selected goals with probability $\epsilon$, which is defined as a function that is related to test success rate $p_s$:
\begin{equation}
\epsilon=\alpha e^{-2p_s} ,
\end{equation}
where $\alpha \in (0,1)$ is a hyperparameter. 
Intuitively, when the test success rate is low, reaching the original goal is beyond the agent's ability, so we encourage agents to explore with the curriculum-based exploration goals. As the test success rate is  increasing, the agent learns to reach more and more original goals. Under this circumstance, we gradually reduce the portion of the relabeled data to alleviate the learning bias.


\subsection{Goal and Transition Augmentation}
\label{Augmentation}
We use augmentation techniques to further expand the  boundary  of achieved goals and increase exploration efficiency.

In sparse reward environments with multiple goals,  a  transition gets a non-negative reward when the achieved goal  is within a small radial threshold (a ball) around the desired goal.
As Fig. \ref{illustration} shows, goals that within the circle with radius $\Delta$ (red dot) also satisfy the success condition for hindsight replacement. These goals form a set $\mathcal{B}_{near}^{g'}=\{g\in \mathcal{G} | d(g, g')< \Delta, \Delta \leq \delta\}$, where $g'$ is a hindsight goal, $d(\bullet)$ is the distance metric used in reward function (Eq. (\ref{reward_fun})) and $\delta$ is the maximum tolerance to get a non-negative reward.
In our practice, after we select a set of goals $\mathcal{B}_{ag}$, we further randomly sample a batch of goals in $\mathcal{B}_{ag}$. For each sampled goal $g'$, we randomly generate an additional goal from its corresponding $\mathcal{B}_{near}^{g'}$, thus forming a set of augmented goals $\mathcal{B}_{aug}$. Then we utilize the set $\mathcal{B}_{ag}\cup \mathcal{B}_{aug}$ to further sample exploration goals following Eq. (\ref{normalize_entropy}). The augmented goals could improve the robustness against the small perturbations. 

For goal-conditioned RL, regardless of the goal being pursued, $(s,a,s')$ transitions are unbiased samples from the environment dynamics. As a result, an agent is free to pair transitions with any goal and corresponding reward. To provide more instructive samples, we propose to utilize the selected goals $\mathcal{B}_{ag}\cup \mathcal{B}_{aug} $ to augment sampled transitions. The transition augmentation technique has two significant advantages: 1) increasing the diversity of learning goals at early exploration; and 2) learning more information of better generalization to desired goals.

Accompanied by the above two augmentation techniques, the agent is capable to achieve more efficient exploration and better generalization.


\section{Experiments and Results}

In this section, we first introduce the simulation environments of  robotic manipulation tasks. Then we present performance results of our method and other baselines as well as analyzing the results. Finally, we conduct ablation experiments to confirm the effectiveness of our method. 

\begin{figure*}
\centering
	\begin{subfigure}[b]{0.18\textwidth}
        \includegraphics[height=2cm,width=3cm]{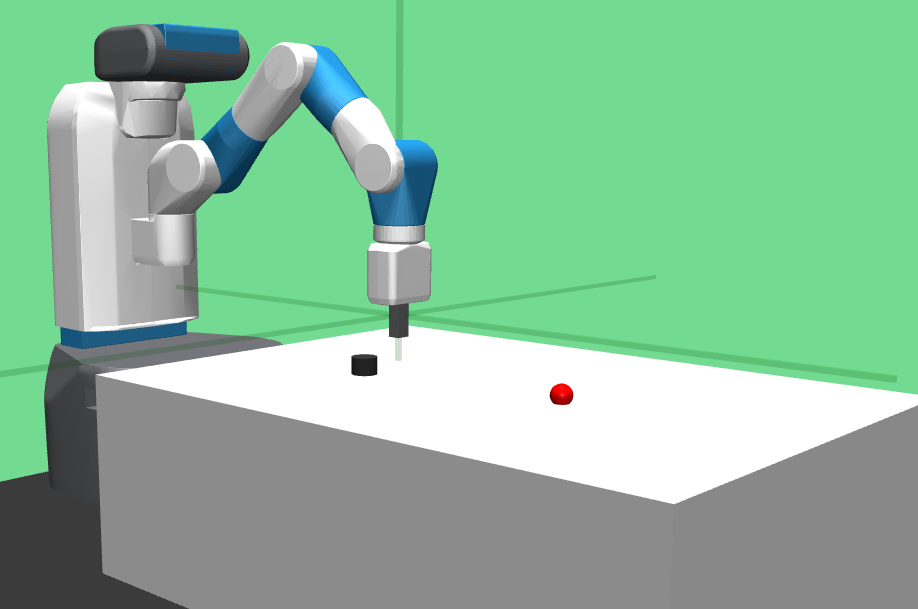}
        \caption{}
	\end{subfigure}
    \hfill
    \begin{subfigure}[b]{0.18\textwidth}
        \includegraphics[height=2cm,width=3cm]{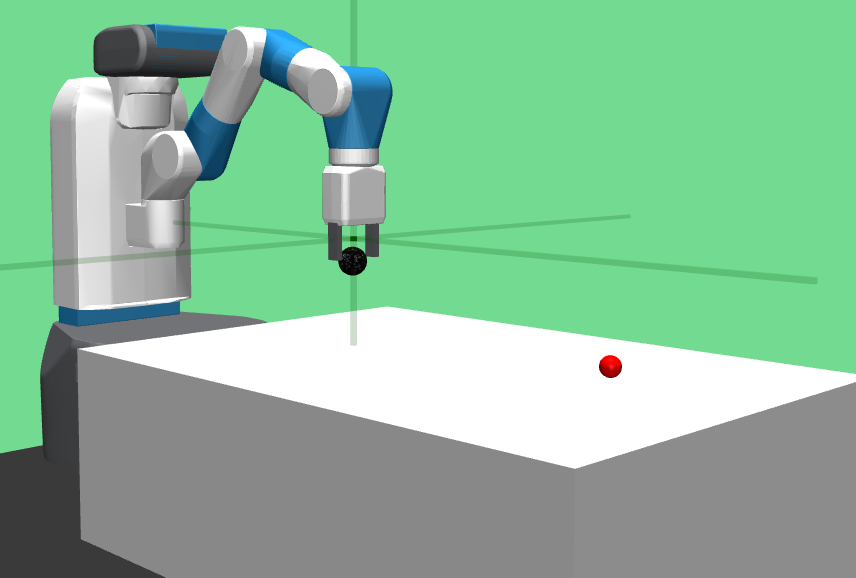}
        \caption{}
	\end{subfigure}
    \hfill
    \begin{subfigure}[b]{0.18\textwidth}
    \includegraphics[height=2cm,width=3cm]{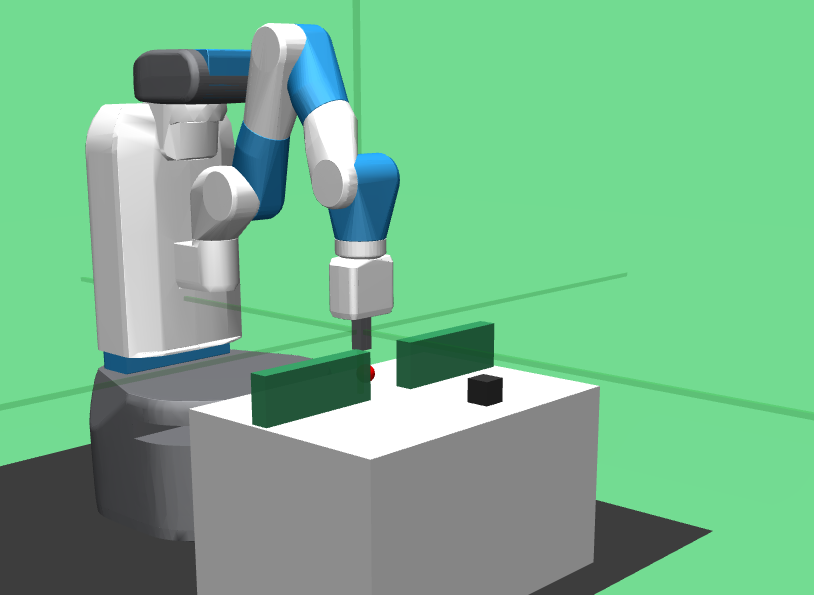}
    \caption{}
    \end{subfigure}
    \hfill
    \begin{subfigure}[b]{0.18\textwidth}
    \includegraphics[height=2cm,width=3cm]{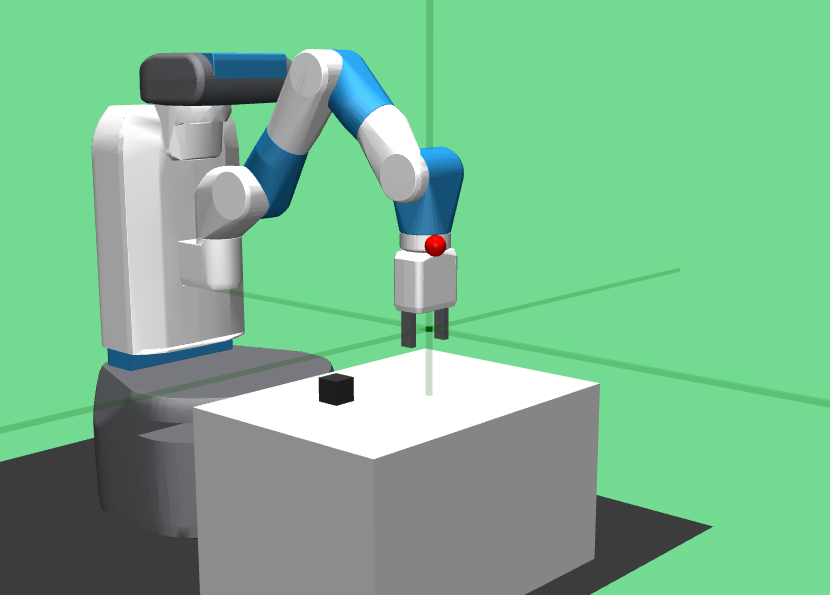}
    \caption{}
    \end{subfigure}
    \hfill
    \begin{subfigure}[b]{0.18\textwidth}
    \includegraphics[height=2cm,width=3cm]{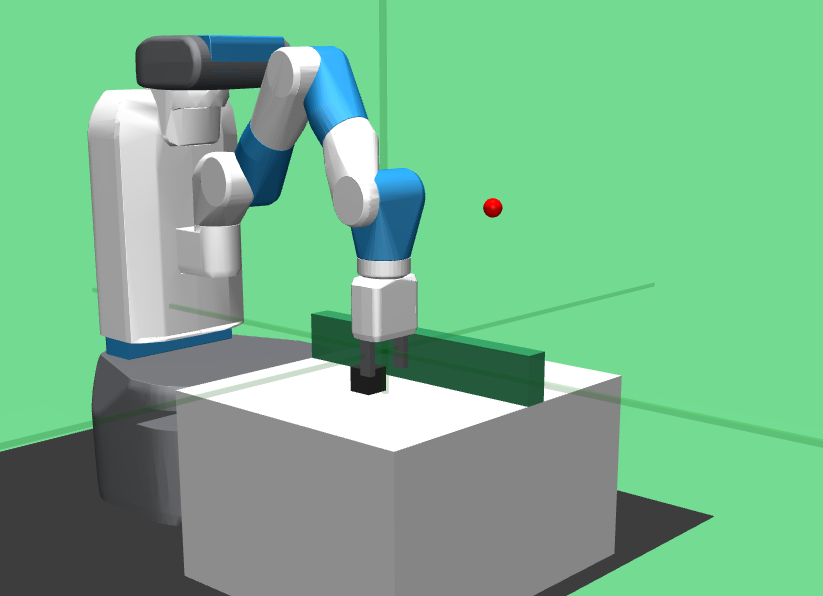}
    \caption{}
    \end{subfigure}
    \hfill

\caption{Five challenging manipulation tasks with sparse reward settings. \textbf{(a)} \textit{FetchSlide-v1}; \textbf{(b)} \textit{FetchThrowBall-v1}; \textbf{(c)} \textit{FetchPush-Obstacle}; \textbf{(d)} \textit{FetchPnP-InAir}; \textbf{(e)}  \textit{FetchPnP-Obstacle}.
}
\hfill
\end{figure*}

\begin{figure*}
\centering
	\begin{subfigure}[b]{0.3\textwidth}
    \centering
        \includegraphics[height=3cm,width=4.5cm]{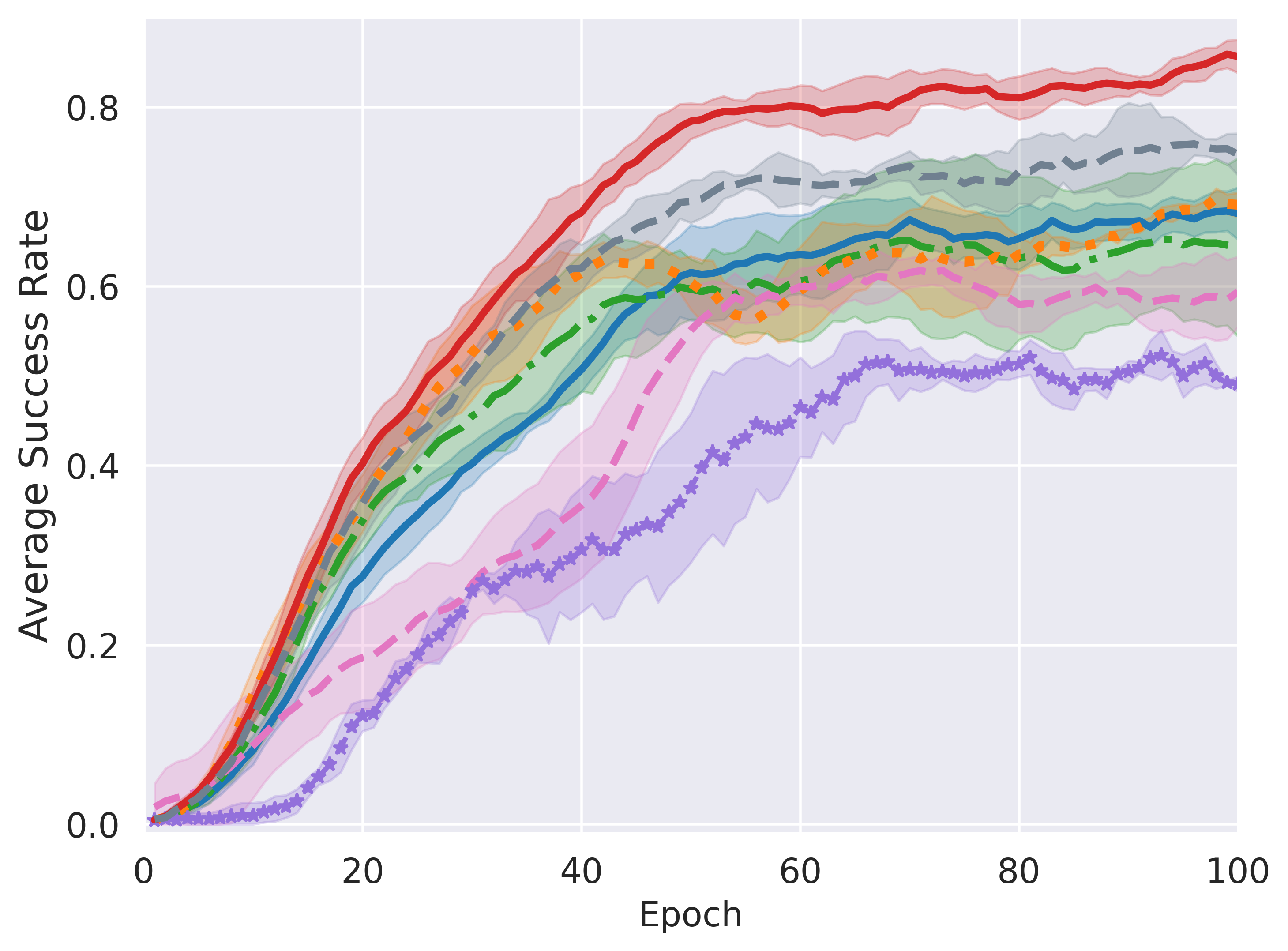}
        \caption{FetchSlide-v1}
        \label{fetchslide}
	\end{subfigure}
    \begin{subfigure}[b]{0.3\textwidth}
    \centering
        \includegraphics[height=3cm,width=4.5cm]{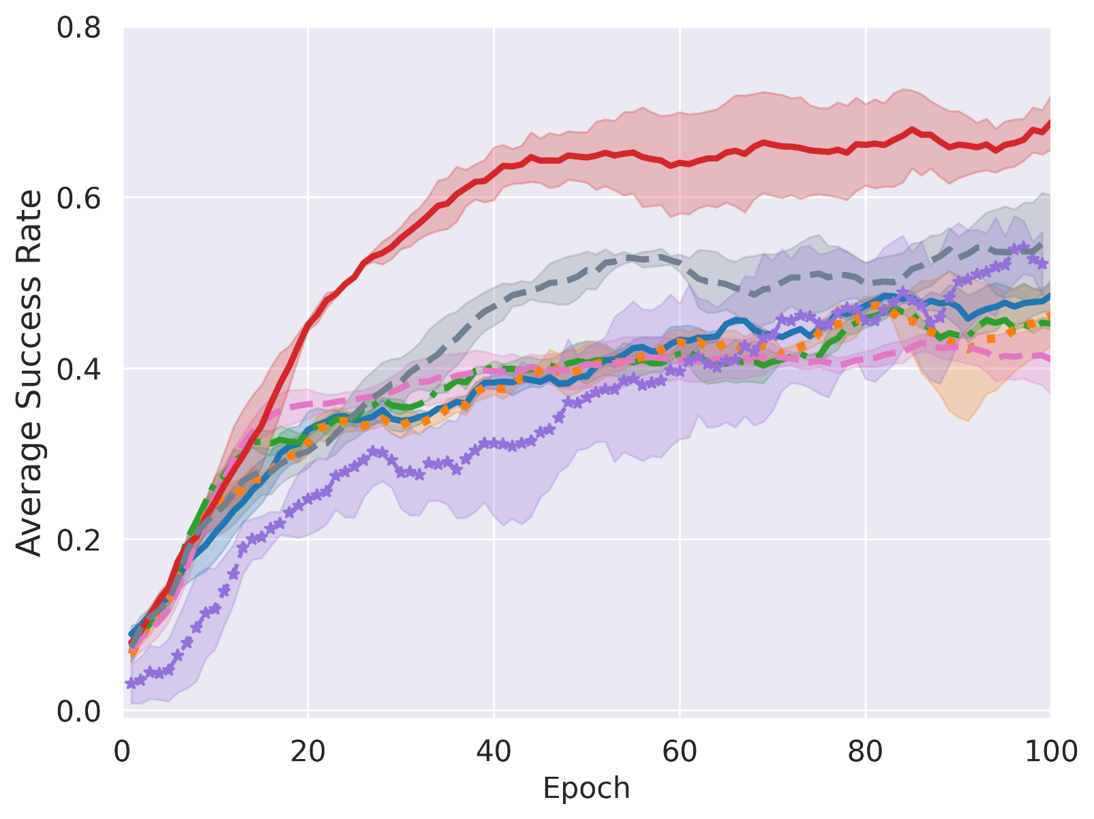}
        \caption{FetchThrowBall-v1}
        \label{fetchthrowball}
	\end{subfigure}
    \begin{subfigure}[b]{0.3\textwidth}
    \centering
    \includegraphics[height=3cm,width=4.5cm]{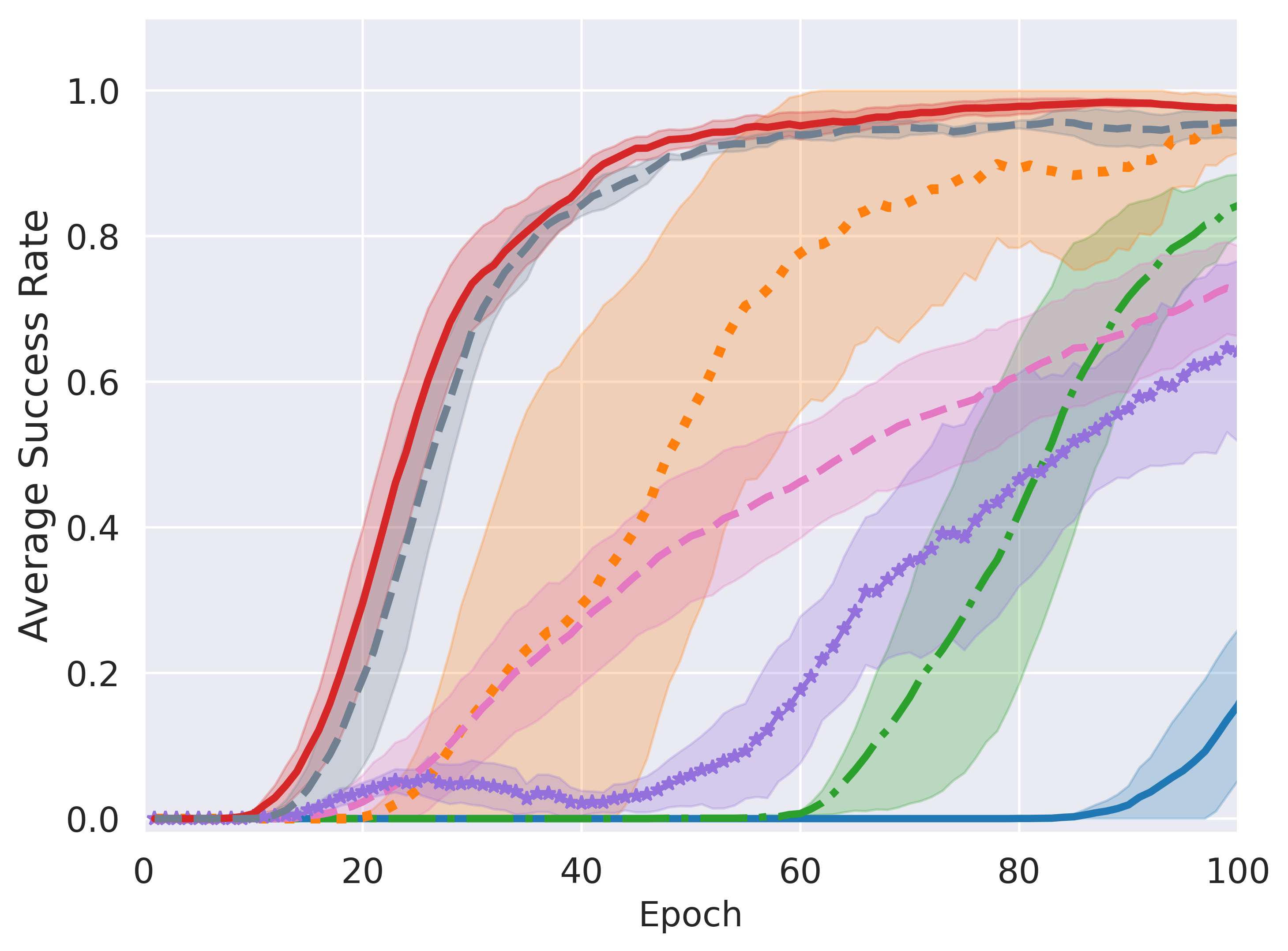}
    \caption{FetchPush-Obstacle}
    \label{fetchpush}
    \end{subfigure}
   
    \begin{subfigure}[b]{0.3\textwidth}
    \centering
    \includegraphics[height=3cm,width=4.5cm]{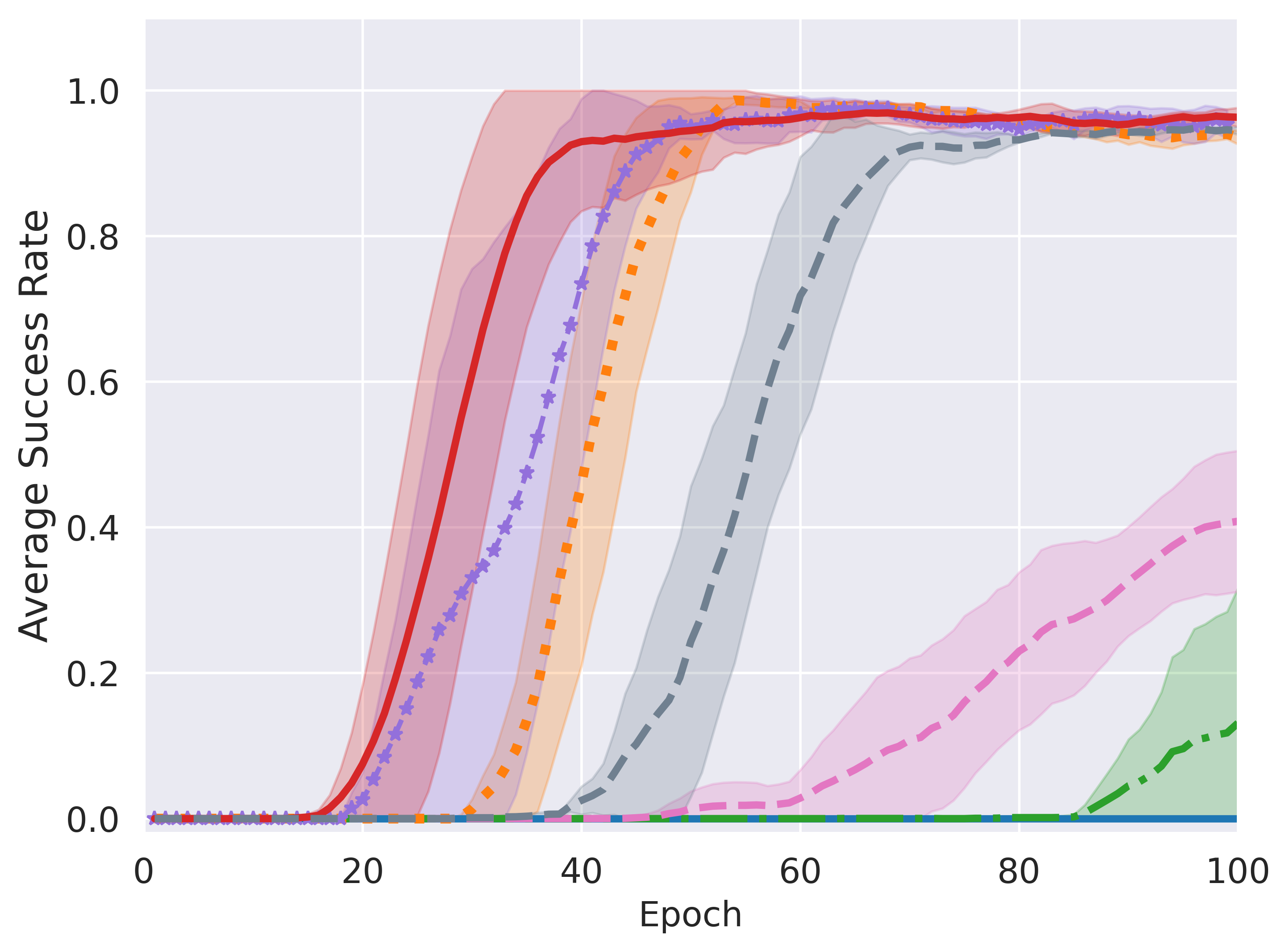}
    \caption{FetchPnP-InAir}
    \label{fetchpnpinair}
    \end{subfigure}
    \begin{subfigure}[b]{0.3\textwidth}
    \centering
    \includegraphics[height=3cm,width=4.5cm]{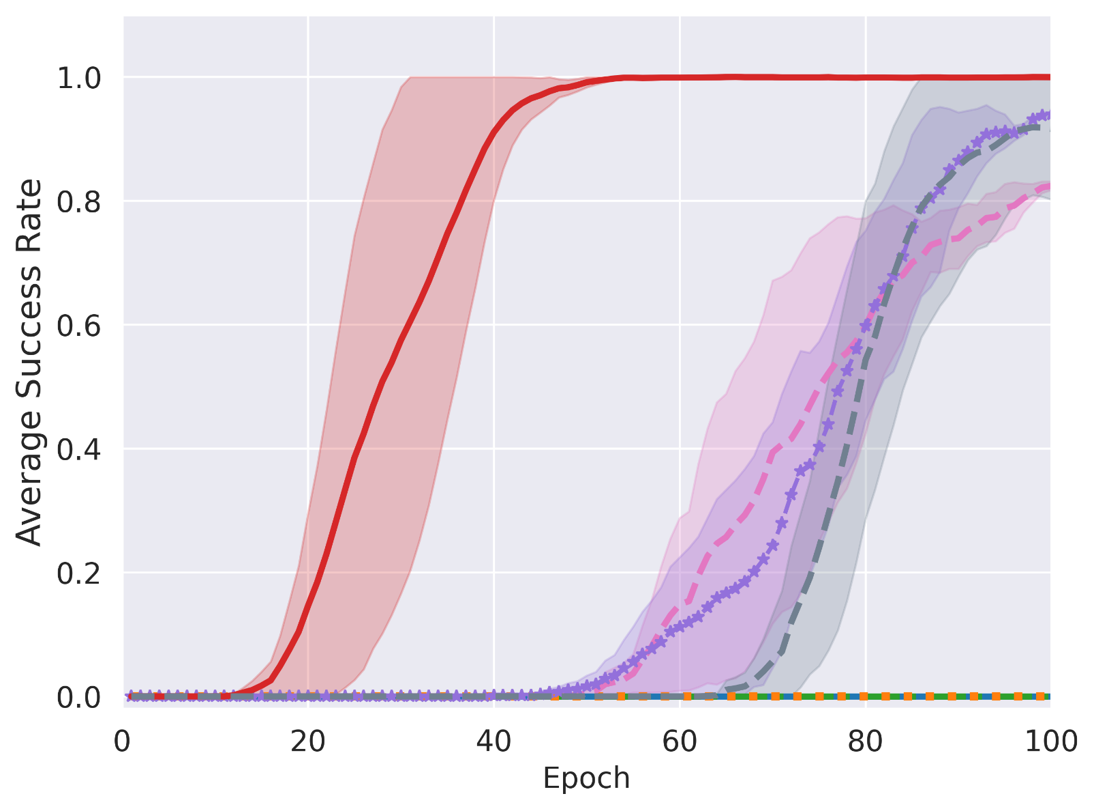}
    \caption{FetchPnP-Obstacle}
    \label{fetchpnpobstacle}
    \end{subfigure}
    \begin{subfigure}[b]{0.3\textwidth}
    \centering
    \includegraphics[height=3cm]{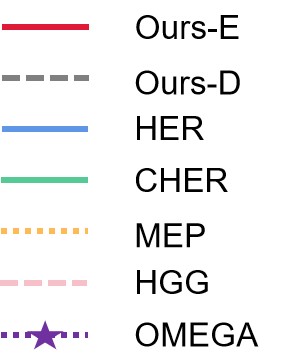}
    \caption{Legend}
    \end{subfigure}
\caption{Learning curves of our method compared with baselines on five challenging robotic manipulation tasks. The solid line indicates the average success rate and the shaded area is the standard error over 5 different random seeds.}
\label{performance}

\end{figure*}

\subsection{Simulation Environments }
The experimental environments are all goal-conditioned tasks with sparse reward settings and simulated with MuJoCo physics engine~\cite{mujoco}. The robot we use in experiments is a 7-DOF (degrees of freedom) Fetch robotic arm with a two-fingered parallel gripper. The action space is 4-dimensional, with 3-dimensional end-effector velocities in $x, y, z$ direction, and 1-dimensional grasping indicator. The goal space is 3-dimensional indicating the Cartesian
coordinate in working space. The state is 25-dimensional for all environments. We evaluate our method on several variants of the standard benchmarks~\cite{Plappert2018MultiGoalRL}  proposed by Open AI gym~\cite{Gym}.
\begin{itemize}
\item \textbf{Sliding} (Fig. \ref{fetchslide}) The robot arm is required to contact the object with enough force and slide the object to a goal position on the table.
\item \textbf{Throwing} (Fig. \ref{fetchthrowball}) Similar to the \textit{Sliding} task, but the robot arm needs to throw a rubber ball to a goal position outside the robot's workspace. The rubber ball is initialized between the two-fingered parallel gripper.
\item \textbf{Pushing} (Fig. \ref{fetchpush}) The robot arm aims to push an object to a position which is far away from initial position. We add two brick-like obstacles to separate the target position and initial position of the object  based on the standard benchmark task \textit{FetchPush-v1}. 
\item \textbf{Pick and Place} We extend the \textit{FetchPickAndPlace-v1} task by increasing the height of desired goals and adding an obstacle. In \textit{FetchPnP-InAir} (Fig. \ref{fetchpnpinair}) task, the heights of desired goals are  between 0.2$m$ and 0.45$m$ above the table, which is the same setting with \cite{Pitis2020MaximumEG}. In \textit{FetchPnP-Obstacle} task (Fig. \ref{fetchpnpobstacle}), the obstacle is added to separate the object initial position and desired goal distribution.
\end{itemize}

\subsection{Training Setting}
We evaluate our method with five baseline algorithms: HER~\cite{HER}, CHER~\cite{CHER}, MEP~\cite{MEP}, HGG~\cite{HGG} and OMEGA~\cite{Pitis2020MaximumEG}. Moreover, to verify the effectiveness of learning uncertain areas of desired goals, we compare our method that uses Eq. (\ref{normalize_entropy}) to select top-ranking goals following entropy prioritization (called Ours\_E) with the one that uses Eq. (\ref{density}) (called Ours\_D).
All the algorithms are combined with DDPG\cite{DDPG} framework. We implement HER, CHER, MEP, HGG  with minor modifications based on the official codebase and unify these baseline algorithms under the same framework. For OMEGA, we use the author's code.  

In all experiments, we use 16 CPU cores to train the agent and each core generates experiences by using 2 parallel rollouts with MPI for synchronization. For OMEGA, we create 16 independent environments and keep the same timesteps  and training frequency as other algorithms. We use the same set of training hyperparameters for HER and our method across all environments except for some specific variables. During training, each epoch generates 16$\times$2$\times$50=1600 episodes. After each epoch, the test success rate is computed over 16$\times$10=160 episodes. We observed that the performance of the proposed approach is robust by choosing explore ratio $\alpha=0.2$ for \textit{FetchSlide-v1} and $\alpha=0.5$ for the other four environments. The size of the selected goal set $\mathcal{B}_{ag}$  and augmented goal set $\mathcal{B}_{aug}$  are 256 and 128 respectively. As for exploration goals, we keep the same size as the number of episodes  in each epoch and select the top 100  from the set $\mathcal{B}_{ag}\cup \mathcal{B}_{aug} $ according to the entropy prioritization. Besides, we choose radius $\Delta=3cm$ for goal augmentation and keep the augmented transition batch size 128 across all environments. The desired and achieved goal buffer size is 1000 and the transition buffer size is $10^6$. 

\begin{figure*}
\centering
	\begin{subfigure}[b]{0.25\textwidth}
    \centering
        \includegraphics[height=3cm,width=4.5cm]{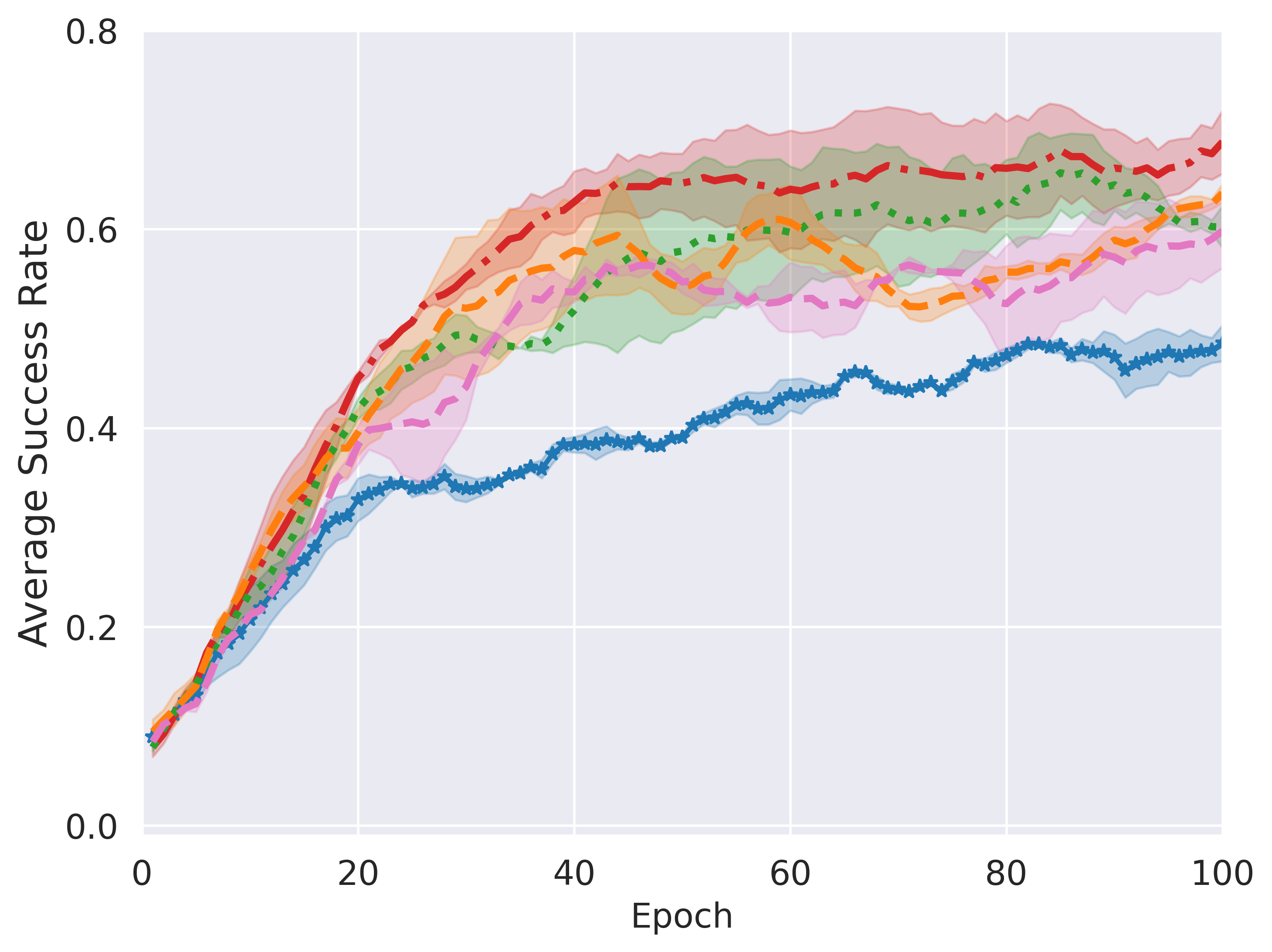}
        \caption{FetchThrowBall-v1}
        \label{ablatiion_fetchthrow}
	\end{subfigure}
    \hfill
    \begin{subfigure}[b]{0.25\textwidth}
    \centering
        \includegraphics[height=3cm,width=4.5cm]{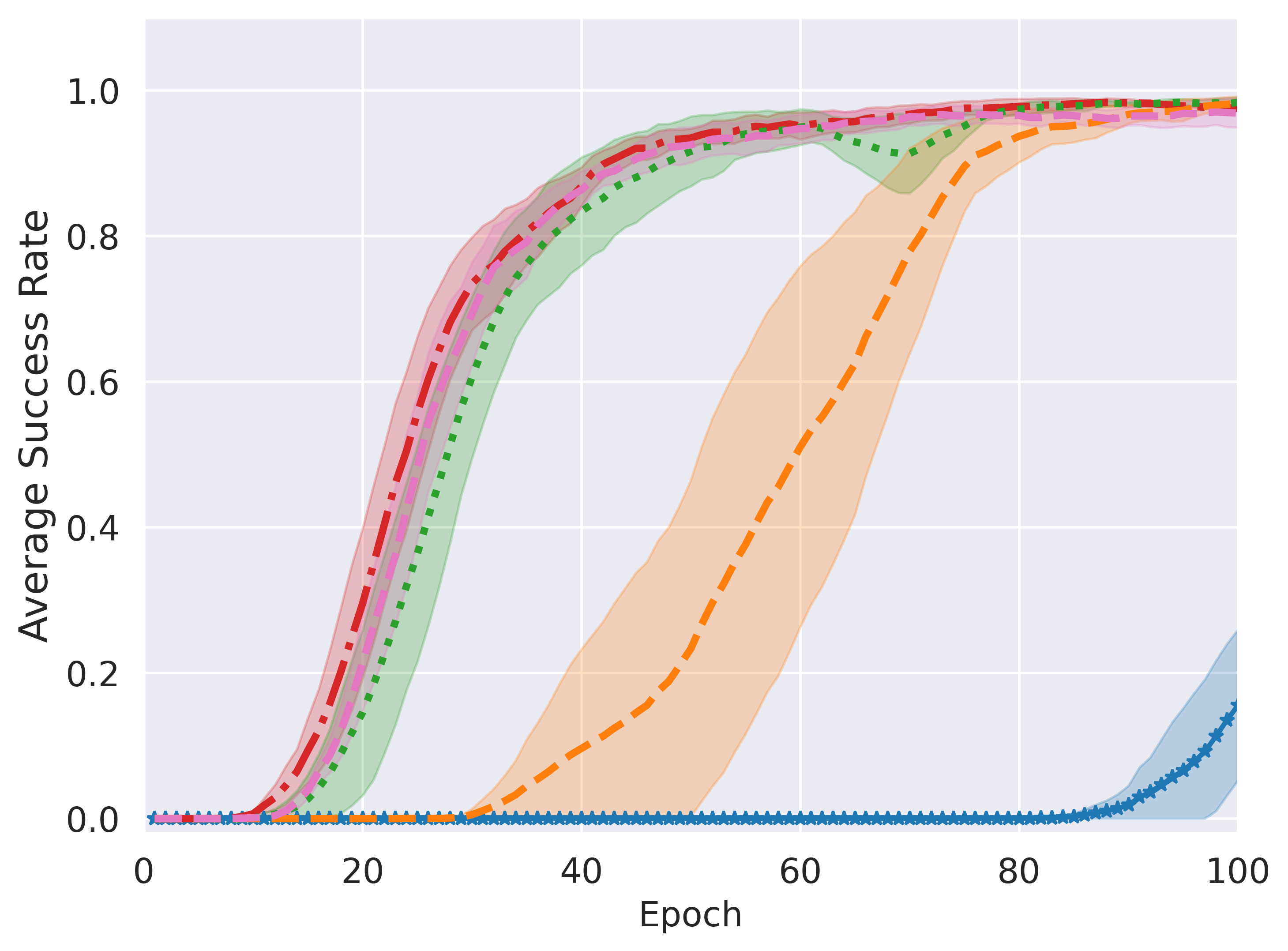}
        \caption{FetchPush-Obstacle}
        \label{ablation_fetchpushobstacle}
	\end{subfigure}
    \hfill
    \begin{subfigure}[b]{0.25\textwidth}
    \centering
    \includegraphics[height=3cm,width=4.5cm]{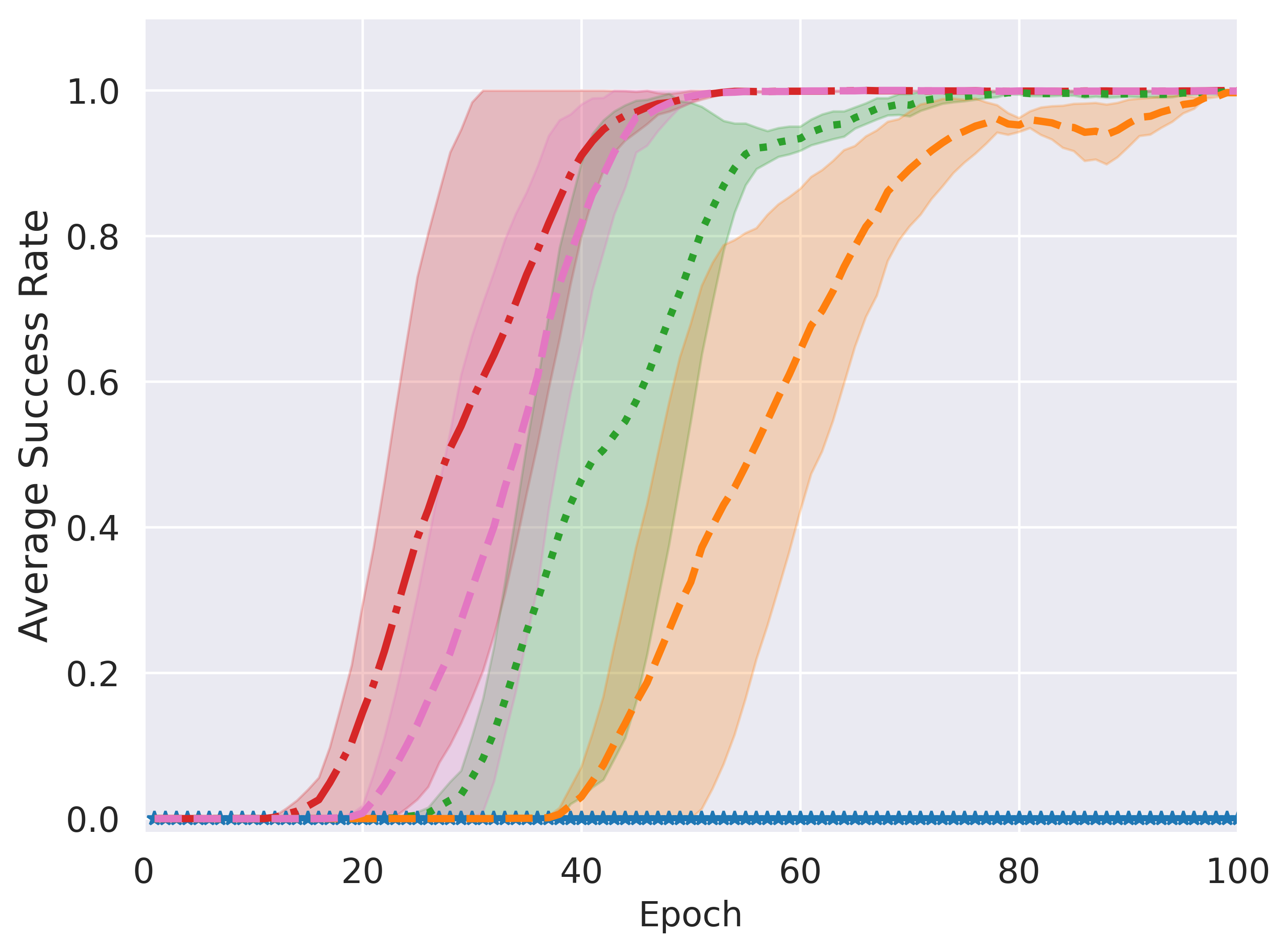}
    \caption{FetchPnP-Obstacle}
    \label{ablation_fetchpnp}
    \end{subfigure}
    \hfill
      \begin{subfigure}[b]{0.2\textwidth}
    \centering
    \includegraphics[height=3cm]{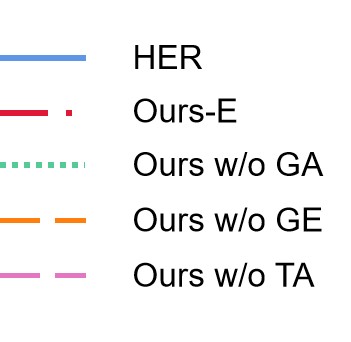}
    \caption{Legend}
    \label{ablation_legend}
    \end{subfigure}
    \hfill
\caption{Ablation study on different components  of our method. Goal exploration (GE), goal augmentation (GA), and transition augmentation (TA). In Fig. \ref{ablation_legend}, w/o represents without.}
\label{ablation_figure}
\end{figure*}

\begin{figure}[h]
	\centering
    \setlength{\belowcaptionskip}{-0.5cm} 
    \includegraphics[width=0.45\textwidth]{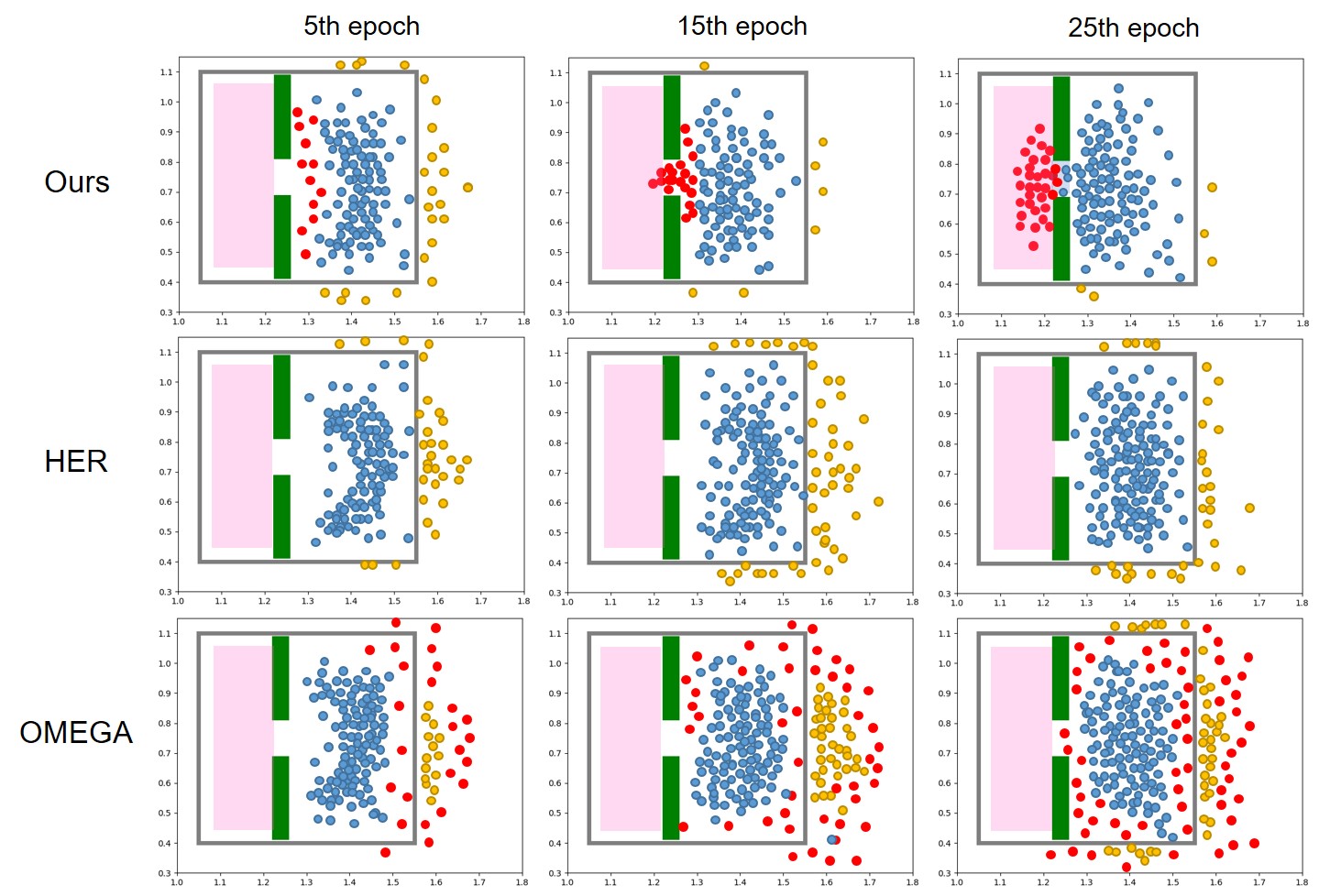}
    \caption{Visualization of hindsight goals on \textit{FetchPush-Obstacle}. The purple area is the region of desired goals. The gray rectangular frame represents the table workspace. The blue dot marks hindsight goal, the red dot marks the exploration goal of our method and  OMEGA, and the orange dot represents the goal below the table. HER does not have red dots because of no exploration goal setting. }
    \label{goal_visualize}
\end{figure}

\subsection{Performance and Analysis}
As shown in Fig. \ref{performance}, our method (Ours-E) achieves state-of-the-art performances in all environments. It can be seen that learning the goals that lie in uncertain areas of desired goal distribution contributes to better performances in all environments compared to choosing high-density goals (Ours-D).

\begin{table}[!t]
\centering
\renewcommand{\arraystretch}{1.3}
\vspace{0.2cm}
\caption{Average training time (hour) for different methods}
\begin{tabular}{@{}lcccccc@{}}
\toprule
     & Ours & HER   & CHER  & MEP  & HGG  & OMEGA \\
\midrule
Time & 3.5h & 3.25h & 13.2h & 5.2h & 4.5h & 8.3h  \\
\bottomrule
\vspace{-0.8cm}
\label{training_time}
\end{tabular}
\end{table}

Besides, we notice that OMEGA has lower sample efficiency than other baselines in \textit{FetchSlide-v1} and \textit{FetchThrowBall-v1}, indicating that maximizes the entropy of  achieved goals may hinder the learning process in those environments that easy to generate unsafe goals (e.g., below the table). However, it's undeniable that entropy-based methods (OMEGA and MEP) contribute to the faster exploration and better generalization than HER. Our method shows significant advantage  in exploration efficiency especially in the most difficult task \textit{FetchPnP-Obstacle}, where HER, CHER, and MEP are failed to solve. Comparing with HGG and OMEGA, our approach achieves over 3$\times$ speedup in  exploration efficiency. 
As  for training time (see Table \ref{training_time}), our method only takes 0.25h longer than HER and has much lower computational complexity than CHER and OMEGA.

To gain a more intuitive understanding of the learned behavior across training, we visualize the achieved goals and exploration goals after the 5th,15th, and 25th epoch (see Fig. \ref{goal_visualize}) during \textit{FetchPush-Obstacle} task training. At the early stage of training, the exploration goals are mainly distributed  on the boundary of the target distribution. Exploring those goals enables agents  approach real targets faster and easier. However, HER always provides the agent completely unreachable goals that are out of the agent's ability and learns  a lot of invalid goals which fall off the table. OMEGA achieves more diverse goals than HER but plenty of goals are in the opposite direction to the desired goals, thus distracting agents from the original purpose. Our method encourages the agent to explore the boundary of  its ability while  getting closer to the real target. Therefore, our method makes agent learning in a curriculum learning manner.

\subsection{Ablation Experiments\label{ablation}}
To measure the effect of each component in our method, we conduct ablation studies on three challenging environments. From Fig. \ref{ablation_figure} we can observe that the learning performance suffers a varying degree of decline without either component of the proposed method, but still achieves better than HER.
It can be seen that transition augmentation is most important for \textit{FetchThrowBall-v1}, indicating that learning goals within  the uncertain area of  desired distribution can further improve the performance. While in the rest of the hard exploration environments, exploration efficiency is severely reduced without setting exploration goals. In addition, goal augmentation brings up an obvious advance in exploration when cooperating with the other two techniques. The synergy between the three components significantly contributes to better exploration efficiency and converged performance.

\begin{figure}
	\centering
    \begin{subfigure}[b]{0.22\textwidth}
    \includegraphics[height=3cm]{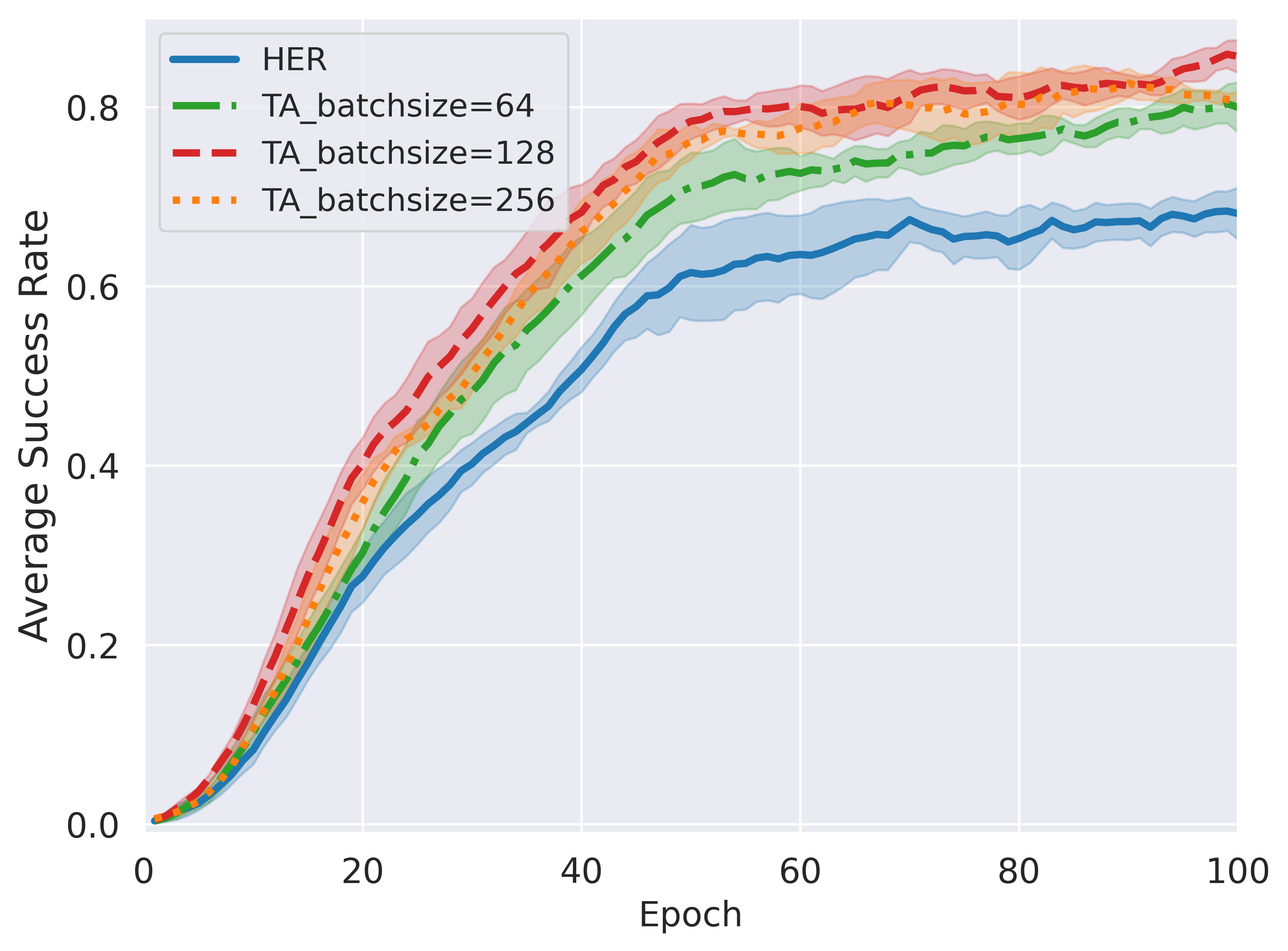}
    \caption{}
    \label{ablation_behbatch}
    \end{subfigure}
    \hfill
    \begin{subfigure}[b]{0.22\textwidth}
    \includegraphics[height=3cm]{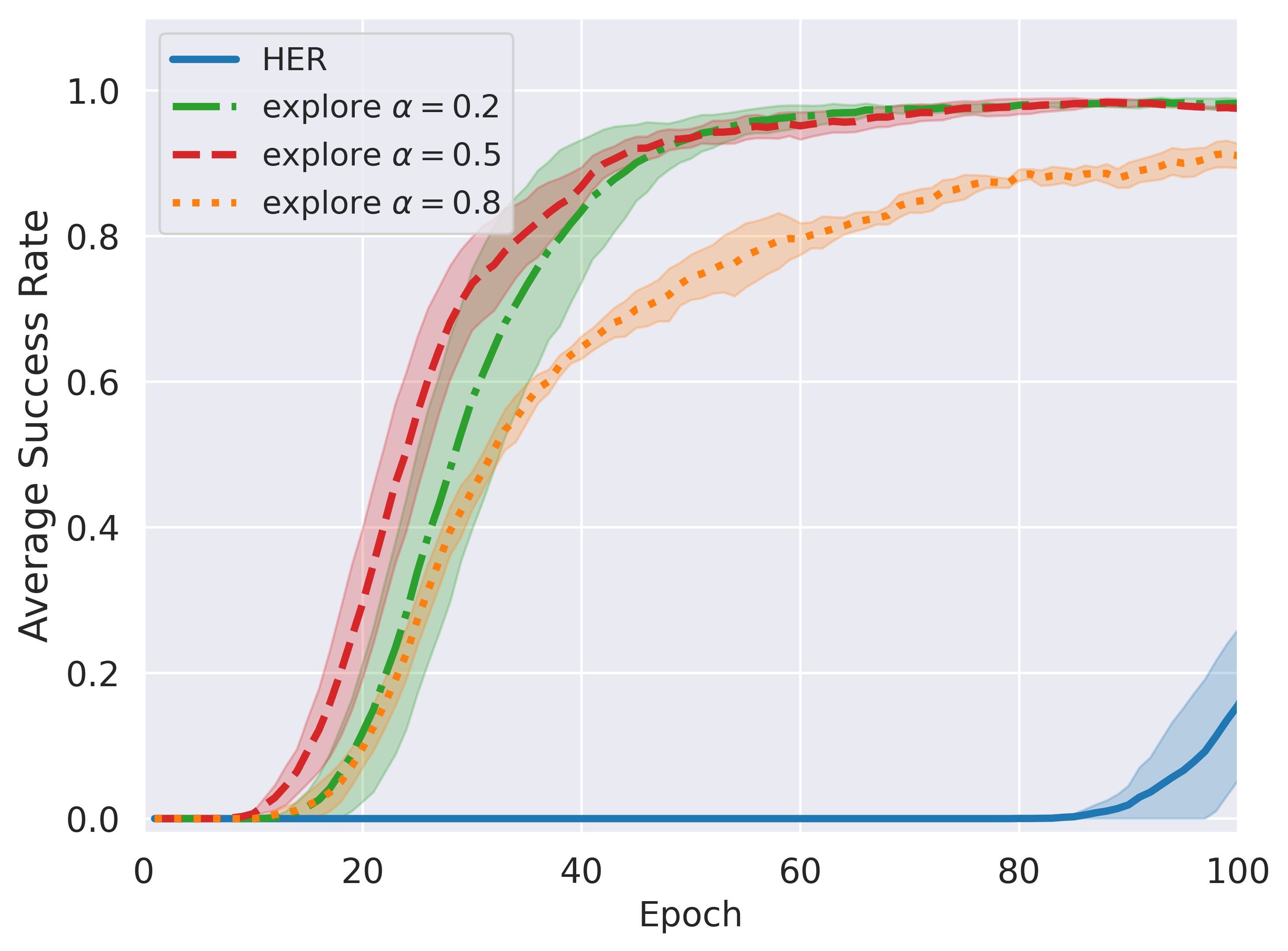}
    \caption{}
    \label{ablation_alpha}
    \end{subfigure}
	\hfill
    \setlength{\belowcaptionskip}{-0.5cm} 
    \caption{Parameters study.\textbf{(a)} Different batch size of transition augmentation on \textit{FetchSlide-v1}; \textbf{(b)} Different explore ratio $\alpha$ on \textit{FetchPush-Obstacle}.}
    \label{parameter_study}
\end{figure}

We also conduct parameter studies (see Fig. \ref{parameter_study}) on the batch size of augmented transitions and explore ratio $\alpha$. It can be seen from Fig. \ref{ablation_behbatch} that less augmented transitions hinder the performance slightly in \textit{FetchSlide-v1}, but still much better than HER. As shown in Fig. \ref{ablation_alpha}, explore ratio $\alpha$ influences the exploration efficiency and performance slightly. A larger explore ratio ($\alpha=0.8$) distracts the agent's attention from learning the original task and decreases the performance. A proper explore ratio ($\alpha=0.5$) balances the learning of local information and original tasks, thus achieving better exploration and performance.

\section{Conclusion}
In this work, we propose a density-based exploration goal selection mechanism to guide goal-conditioned  agents to explore efficiently towards  the frontier of desired distribution. Cooperating with two simple but effective techniques: goal augmentation and transition augmentation, the agent is capable to  generalize hindsight goal distribution better. We evaluate our method in five challenging sparse reward robotic environments and demonstrate  substantial improvements in performance and sample efficiency compared to HER and other standard RL algorithms. Through further analysis, we demonstrate that each component of our method contributes a positive effect on exploration efficiency. Future research could concentrate on extension (e.g., image-based tasks), improvement, and real-world deployment  of our method.


\bibliographystyle{ieeetr}


\end{document}